\documentclass[letterpaper, 10 pt, journal, twoside]{IEEEtran}

\IEEEoverridecommandlockouts                              

\usepackage[bookmarks=true]{hyperref}

\usepackage{caption}
\usepackage{subcaption}
\usepackage{physics}
\usepackage{xcolor}

\usepackage{amsmath,amsfonts}
\usepackage{algorithmic}
\usepackage{algorithm}
\usepackage{array}
\usepackage{textcomp}
\usepackage{stfloats}
\usepackage{url}
\usepackage{verbatim}
\usepackage{graphicx}
\usepackage{cite}
\usepackage{hyperref} 
\usepackage{color,soul}
\definecolor{red}{rgb}{1.00,0.00,0.00}
\definecolor{blue}{rgb}{0.00,0.00,1.00}
\definecolor{green}{rgb}{0.30, 0.50,0.00}
\definecolor{cyan}{RGB}{0, 128, 255}
\definecolor{orange}{rgb}{1.00,0.50,0.00}
\newcommand{\cred}[1] {\textcolor{red}{#1}}
\newcommand{\cblue}[1] {\textcolor{blue}{#1}}
\newcommand{\cgreen}[1] {\textcolor{green}{#1}}
\newcommand{\ccyan}[1] {\textcolor{cyan}{#1}}
\newcommand{\corange}[1]{\textcolor{orange}{#1}}
\usepackage{bm}
\usepackage{booktabs} 
\usepackage{tabularx}  
\usepackage{mwe}
\usepackage{wrapfig}
\usepackage{pifont}

\usepackage{multirow}

\begin{document}

\title{
Learning Dual-Arm Push and Grasp Synergy \\in Dense Clutter}
\author{Yongliang Wang and Hamidreza Kasaei
\thanks{Manuscript received: November 27, 2024; Revised: January 27, 2025; Accepted: March 24, 2025.

This paper was recommended for publication by Editor Júlia Borràs Sol upon evaluation of the Associate Editor and Reviewers’ comments.

\textit{\noindent(Corresponding author: Hamidreza Kasaei})}
\thanks{ The authors are with the Department of Artificial Intelligence, Bernoulli Institute, Faculty of Science and Engineering, University of Groningen, The Netherlands. \scriptsize\texttt{\{yongliang.wang, hamidreza.kasaei\}@rug.nl}} 
\thanks{Digital Object Identifier (DOI): see top of this page.}
}

\maketitle

\begin{abstract}

Robotic grasping in densely cluttered environments is challenging due to scarce collision-free grasp affordances. Non-prehensile actions can increase feasible grasps in cluttered environments, but most research focuses on single-arm rather than dual-arm manipulation. Policies from single-arm systems fail to fully leverage the advantages of dual-arm coordination. We propose a target-oriented hierarchical deep reinforcement learning (DRL) framework that learns dual-arm push-grasp synergy for grasping objects to enhance dexterous manipulation in dense clutter. Our framework maps visual observations to actions via a pre-trained deep learning backbone and a novel CNN-based DRL model, trained with Proximal Policy Optimization (PPO), to develop a dual-arm push-grasp strategy. The backbone enhances feature mapping in densely cluttered environments. A novel fuzzy-based reward function is introduced to accelerate efficient strategy learning. Our system is developed and trained in Isaac Gym and then tested in simulations and on a real robot. Experimental results show that our framework effectively maps visual data to dual push-grasp motions, enabling the dual-arm system to grasp target objects in complex environments. Compared to other methods, our approach generates 6-DoF grasp candidates and enables dual-arm push actions, mimicking human behavior. Results show that our method efficiently completes tasks in densely cluttered environments. \href{https://sites.google.com/view/pg4da/home}{\cblue{https://sites.google.com/view/pg4da/home}}

\end{abstract}


\section{Introduction}
\label{sec: introduction}

\begin{figure}[!htbp]
      \centering
      \includegraphics[width=1.0\linewidth]{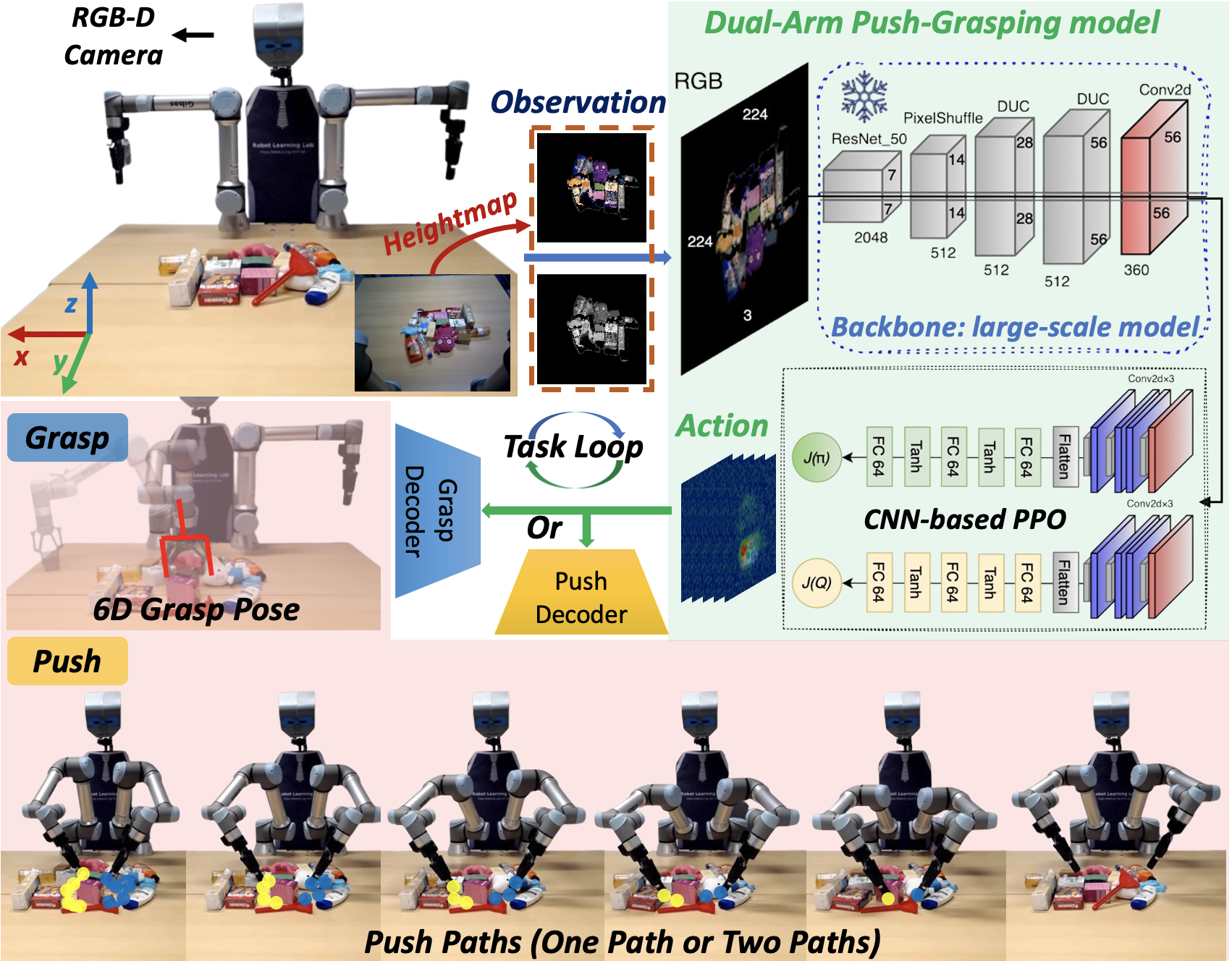}
      \vspace{-4mm}
      \caption{An example of a target-oriented grasping task where the robot aims to grasp a green object in dense clutter. Tight packing around the target requires synergy between pushing and grasping actions. Our system maps RGBD images to actions, decides on Push or Grasp based on the current state, and plans suitable push actions (dual or single paths) to isolate the object before executing a stable 6-DoF grasp to complete the task.}
      \label{fig: problem}
\end{figure}

\IEEEPARstart{O}{bject} grasping is crucial in robotic manipulation as it is fundamental for numerous complex tasks. In many applications, robots operate in cluttered environments where identifying feasible, collision-free grasping affordances is challenging due to spatial constraints. Inspired by human behavior, combining pushing and grasping provides an effective solution in cluttered environments. By pushing aside surrounding objects, space is created around the target, enabling successful grasping by the robot. Most research focuses on developing manipulation policies using single-arm robots to replicate this skill. However, in densely cluttered environments, a single arm often requires significantly more actions than a dual-arm robot to efficiently declutter the scene.



Grasping tasks in robotics are generally categorized into two main types: \textbf{target-agnostic} and \textbf{target-oriented}. For target-agnostic tasks, \cite{zeng2018learning} employs two parallel DRL networks to learn the policy equally. Building on transformers and CNNs, \cite{yu2023novel} introduces a new push-grasp detection network, combining a grasping network with a vision transformer (ViT)-based object position prediction network and a pushing transformer network. For target-oriented tasks, \cite{xu2021efficient} proposes a target-oriented hierarchical reinforcement learning approach with high sample efficiency to learn a push-grasping policy for specific object retrieval in clutter. Additionally, \cite{li2022learning} investigates target-oriented push-grasping synergy in cluttered environments using reinforcement learning (RL) without relying on object detection and segmentation. \cite{ren2023learning} introduces a bifunctional push-grasping strategy for both target-agnostic and target-oriented tasks. Their method combines pushing and grasping to efficiently pick up all objects or specific target objects, depending on the task requirements. However, most methods focus on top-down grasping with push actions limited to predefined straight-line distances. While these methods work well in clutter, they are less effective in dense clutter environments.

Several studies use dual-arm robots to develop policies beyond push-grasping. For instance, \cite{driess2020deep} introduces a deep convolutional recurrent neural network to predict action sequences for task and motion planning (TAMP) from initial scene images. \cite{yu2023coarse} presents a coarse-to-fine framework combining global planning and local control for dual-arm manipulation of deformable linear objects (DLOs), enabling precise positioning and collision avoidance. \cite{kim2024goal} proposes a target-conditioned dual-action deep imitation learning (DIL) method to learn dexterous manipulation skills from human demonstrations. Similarly, \cite{cui2024task} develops a DRL framework for closed-chain manipulation with dual-arm systems. \cite{gao2024toward} presents a pipeline for dual-arm TAMP. While these studies focus on dual-arm motion planning and manipulation, they do not address self-supervised learning for dexterous manipulation. 

In this paper, we propose a self-supervised DRL framework for dual-arm robots to learn coordinated push-grasp actions in target-oriented tasks within densely cluttered environments. As shown in Fig.~\ref{fig: problem}, Our model processes RGB-D images to generate adaptive dual-arm push actions, enabling the grasping of target objects that are otherwise challenging to grasp directly. Unlike existing research, we treat the dual-arm system as a learning agent and frame the task as a target-conditioned hierarchical reinforcement learning problem, training the model from scratch. Our framework uses a large-scale DL model as the backbone for multi-degree-of-freedom (multi-DoF) grasping, integrating a custom CNN-based RL model, trained with PPO, to learn the push-grasping strategy. We introduce a novel fuzzy-based reward function to accelerate efficient strategy learning. Our system is developed and trained in Isaac Gym and tested in simulations and on a real robot. Experiments in simulated and real-world environments show that our system outperforms others in target-oriented tasks, achieving higher task completion and grasping success rates with fewer steps. It transitions seamlessly from simulation to real-world application without additional data collection or fine-tuning. In summary, the contributions are:

\begin{itemize}
    \item We propose an adaptive push action generation method that samples from the learned feature map, projects connected pixel points into 3D space, and smooths the 3D trajectory with the Savitzky-Golay filter. Unlike previous simplistic push methods, our approach enables flexible, dual-arm push actions with single or coordinated paths.
    
    \item Our model addresses the limitations of conventional top-down grasp poses in push-grasping tasks by outputting 6-DoF grasp poses, enabling more adaptable and precise grasping in densely cluttered environments.
    
    \item To efficiently learn dual-arm push-grasp synergy, we design a fuzzy reward function to guide the model by evaluating action validity and push or grasp suitability.
    
    \item To accelerate learning, we develop two robotic system versions in Isaac Gym: a training version with simplified grippers and no kinematic constraints, and a full system with manipulators for testing. We evaluate our framework in simulated dense clutter and real-world scenarios without fine-tuning, confirming its effectiveness.
\end{itemize}

\section{Related work}
\label{sec: related_work}

\subsection{Target-driven Object Grasping}
\label{subsec: target-driven}

Robotic grasping methods have been extensively studied over the past decades, broadly divided into model-based and learning-based approaches \cite{bohg2013data, yang2024attribute, newbury2023deep, morrison2020learning}. They can also be categorized by task objectives into target-agnostic \cite{mousavian20196, song2020grasping, fang2023anygrasp} and target-driven \cite{xu2023joint, lu2023vl} methods.

Target-agnostic grasping methods use deep neural networks to learn either top-down or 6-DoF grasp poses. For top-down grasping, \cite{mahler2017dex} introduces the grasp quality convolutional neural network (GQ-CNN) to predict grasp success from depth images, while \cite{morrison2018closing, morrison2020learning} proposes the generative grasping convolutional neural network (GG-CNN), predicting grasp quality and pose per pixel. However, top-down poses limit complex tasks, prompting recent 6-DoF developments \cite{mousavian20196, song2020grasping}. \cite{gou2021rgb} improves 7-DoF detection with RGB Matters and introduces AnyGrasp, a robust grasp perception system \cite{fang2023anygrasp}. \cite{liu2024simulating} enhances single-view 6-DoF detection via knowledge distillation from a full-point cloud detector, while \cite{chen2023efficient} presents a 6-DoF local grasp generator with a grid-based strategy, Gaussian encoding, and novel non-uniform anchor sampling.

Target-driven grasping has received less attention and requires a perception module to identify specific objects. \cite{xu2023joint} combines vision, language, and action with object-centric representation for flexible instructions, improving sample efficiency and sim2real transfer. \cite{lu2023vl} presents Visual-Lingual-Grasp (VL-Grasp), an interactive policy integrating language cues, a visual grounding dataset, and a 6-DoF grasp policy. Some methods use prehensile motion-assisted grasping, though most assume isolated objects in scattered scenes. Relying solely on grasp actions is inadequate for dense clutter.

\subsection{Strategy of Push-grasping}
\label{subsec: strategy}

The synergy between pushing and grasping has been studied to rearrange clutter, enabling more effective grasps \cite{pang2023object}. \cite{zeng2018learning} introduced visual pushing for grasping (VPG), a model-free DRL framework that learns joint push-grasp policies through a parallel architecture, launching push-grasp synergy research. \cite{yu2023novel} proposes a vision transformer-based pushing network (PTNet) with a cross-dense fusion network (CDFNet) for precise grasp detection, while \cite{yu2024efficient} presents an end-to-end push-grasp method using EfficientNet-B0 and a cross-fusion module. These methods are target-agnostic. For target-oriented push-grasping, \cite{liu2022ge} introduces GE-Grasp, a framework with diverse action primitives and a generator-evaluator DL architecture for grasping in dense clutter, even with occlusions. Similarly, \cite{cao2024plot} presents PLOT, using a target RGB-D image, object segmentation, feature matching, and self-supervised Q-learning for efficient target grasping. \cite{ren2023learning} employs a bifunctional network with hierarchical RL, supporting both target-agnostic and target-oriented tasks. \cite{chen2023minimal} addresses push-grasping in large clutter, while \cite{gao2024improved} applies it underwater. \cite{zhang2023reinforcement} uses a Variational Autoencoder and PPO to efficiently learn push-grasping for large flat objects like books. \cite{tang2021learning} jointly learns to push and 6-DoF grasp in clutter, while \cite{huang2021visual} combines neural prediction and tree search to optimize push actions efficiently. \cite{bejjani2021occlusion} develops an RL-guided hybrid planner for retrieving occluded objects, while \cite{dogar2010push} and \cite{kiatos2022learning} explore push-grasp for stable multi-finger manipulation in clutter. However, these approaches rely on single manipulators, limiting efficiency in dense clutter and highlighting the value of dual-arm push-grasping.

\subsection{Reinforcement Learning for Manipulation}
\label{subsec: reinforcement}

With the rise of RL in robotic manipulation, various DRL models have been developed for these tasks \cite{yang2023dynamo}. Key RL challenges include designing effective state-action representations and reward mechanisms. In robotic manipulation, Deep Q-learning (DQN) is often used to learn directly from visual information, but it typically falls short in complex tasks requiring sophisticated actions. For example, DQN-based push-grasping methods generate simple push directions with fixed lengths, inefficient in densely cluttered environments \cite{zeng2018learning, xu2021efficient, wu2023learning}. Most research extracts visual features and converts them into vectors as state representations for DRL models \cite{nagato2023probabilistic, wang2024multi}. However, a common approach involves designing parallel networks for different primitive motions, which is resource-intensive and requires extensive parameter tuning. Thus, developing a lightweight model that allows direct learning from visual cues to actions remains an active research area.

In summary, existing DRL models employ various approaches to learning from visual input. Some rely solely on DQN for visual processing, while others preprocess visual data to extract pose vectors, feeding these into RL models to learn motions. Additionally, some methods require multiple models or multi-stage processes. In contrast, our approach streamlines this process, learning directly from RGB images without needing parallel networks to manage different primitive motions.

\section{Method}
\label{sec: method}

\subsection{System Overview}
\label{subsec: system}

\begin{figure*}[!htbp]
      \centering
      \includegraphics[width=1.0\linewidth]{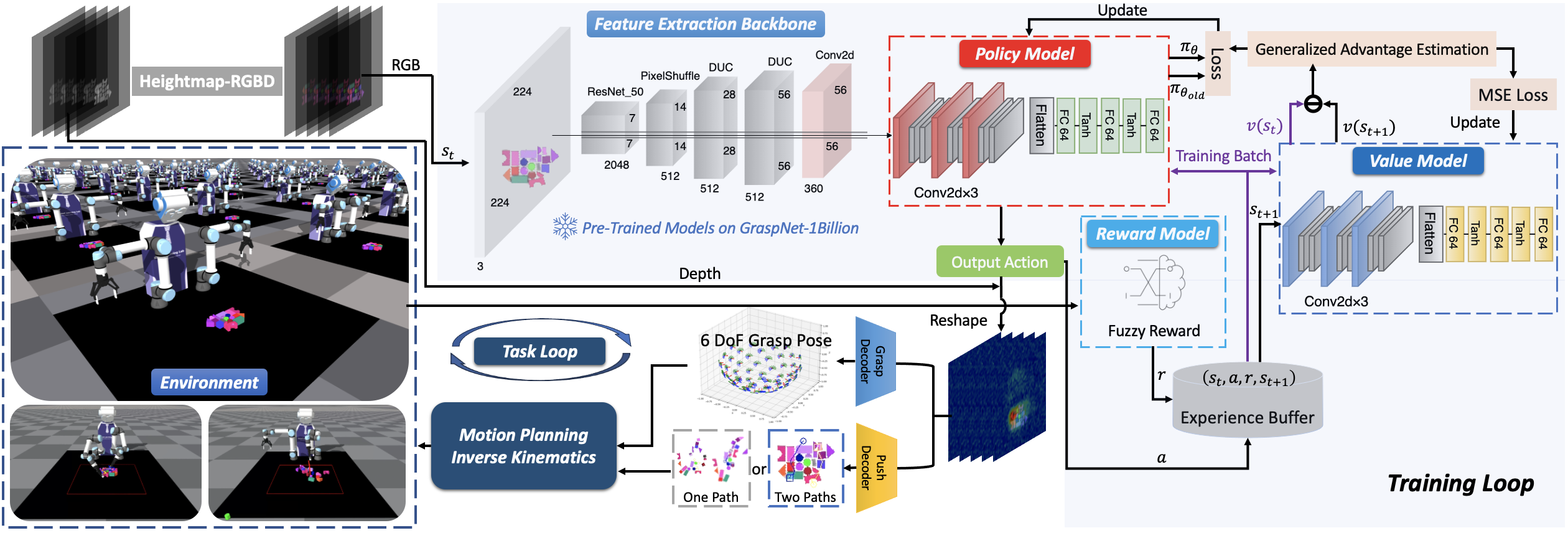}
      \vspace{-4mm}
      \caption{\textbf{Dual-Arm Push-Grasping Learning Framework Overview}: In the Isaac Gym environment, the target object is highlighted in green, and an RGB-D camera integrated with a dual-arm UR5e robot captures images, converting them into top-down height maps. RGB images are input into a grasp network pre-trained on GraspNet-1 Billion, which extracts features fed into a CNN-based RL model trained with PPO. This model produces a feature map, decoded by two motion decoders to generate actions within the environment. A fuzzy reward module provides feedback and guiding training in the light blue area.}
      \label{fig: system_overview}
\end{figure*}

We model the target-oriented dual-arm push-grasping task as a target-conditioned Markov Decision Process (MDP) within a hierarchical reinforcement learning framework, consistent with prior work \cite{zeng2018learning, xu2021efficient, wu2023learning}. Unlike previous approaches that primarily rely on Deep Q-Networks (DQN), our framework enables direct learning from visual inputs to actions. As illustrated in Fig.~\ref{fig: system_overview}, an overhead camera on the dual-arm robot captures RGB-D images of the workspace. Each state \( s_t \) is represented by a color heightmap \( c_t \) and depth heightmap \( d_t \), created by projecting RGB-D images along the gravity axis. This setup allows the agent to learn target-directed behavior directly from raw visual data. The dual-arm push-grasping task is formulated as a hierarchical reinforcement learning problem, decomposed into two sub-tasks. First, the \textbf{Angle-View Network (AVN)} backbone processes RGB images to predict gripper orientations at various image positions. Second, we introduce a specialized \textbf{CNN-based RL model} trained with PPO to generate dual-arm push-grasping actions. Our framework uses RGB channels in the backbone network for feature map generation, while depth images refine and filter the RL model’s actions, enhancing agent decision-making. Unlike current methods, our framework generates 6-DoF grasp candidates rather than top-down grasps. Our learned push actions are also more adaptive, with flexible orientations and pushing lengths, resulting in versatile and efficient strategies that enhance dual-arm manipulation performance.

\begin{figure}[!htbp]
      \centering
      \includegraphics[width=1.0\linewidth]{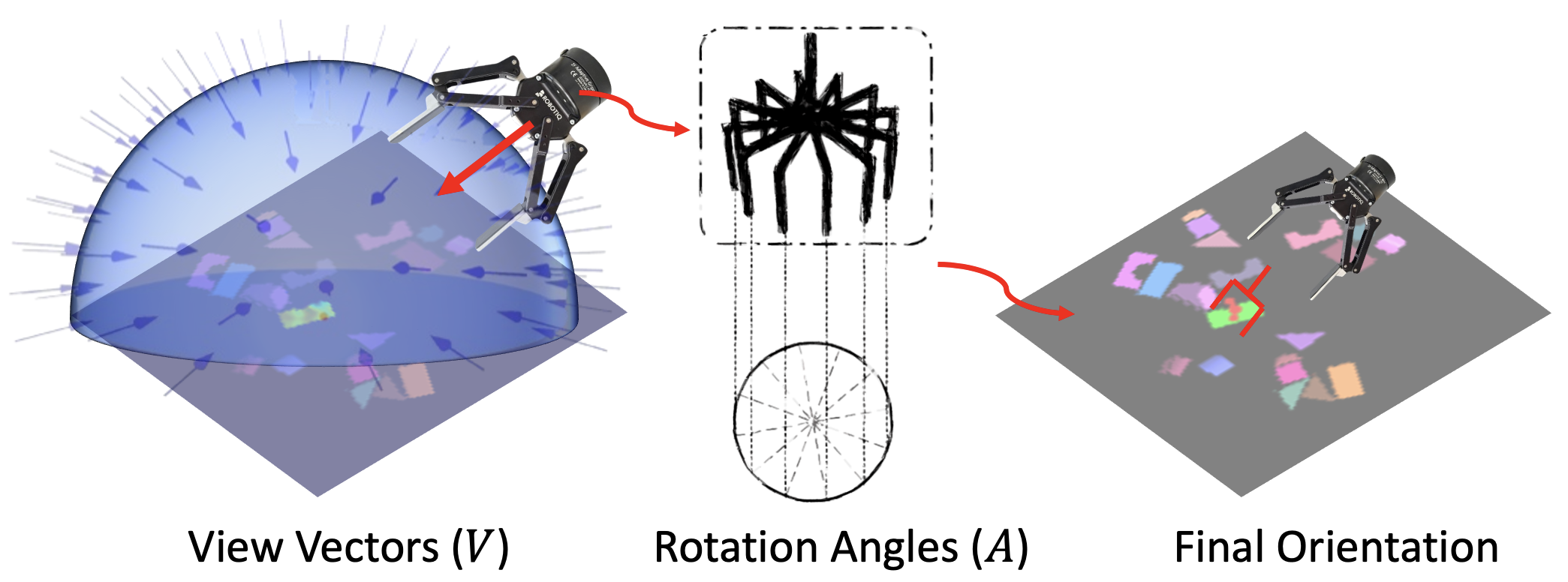}
      \vspace{-4mm}
      \caption{\textbf{From Angle View to Final Orientation}: Gripper orientation for grasping is determined by a view vector and an in-plane rotation angle. In the left section of the figure, $V$ view vectors are uniformly sampled across the upper hemisphere, while in the middle, $A$ in-plane rotation angles are sampled. Here, $V$ and $A$ are set to $60$ and $6$, respectively. The model outputs a value $(0–359)$, which is then decoded to determine the final gripper orientation.}
      \label{fig: va}
\end{figure}

\subsection{Model Architecture}
\label{subsec: model}

\subsubsection{Angle-View Net}
\label{subsubsec: angle}

The AVN predicts pixel-wise gripper rotation configurations. Directly regressing rotation matrices or quaternions is impractical due to multiple viable rotations for stable grasping at each location. Inspired by GraspNet-1Billion, the AVN decomposes gripper orientation into two parts: approach direction and in-plane rotation, framing the task as a multi-class classification problem to better handle rotational variability. As shown in Fig.~\ref{fig: va}, $V$ approach directions are uniformly sampled from the upper hemisphere, and $A$ in-plane rotations are sampled, creating $V \times A$ distinct orientation classes to capture a wide range of possible configurations.

To determine gripper orientations across the image, the AVN divides it into a grid of size $G_H \times G_W$, where $G_H$ and $G_W$ represent the number of grid cells along the vertical and horizontal axes, respectively. For each grid cell, the AVN outputs a 1-dimensional vector with $V \times A$ elements, representing confidence scores for each orientation class within that cell. The final output, termed the Angle-View Heatmap (AVH), is represented as a 3D tensor:

\begin{equation}
    \textbf{AVH} = \mathbb{R}^{(V \times A) \times G_H \times G_W}
    \label{equ: 1}
\end{equation}

\begin{figure}[!htbp]
      \centering
      \includegraphics[width=1.0\linewidth]{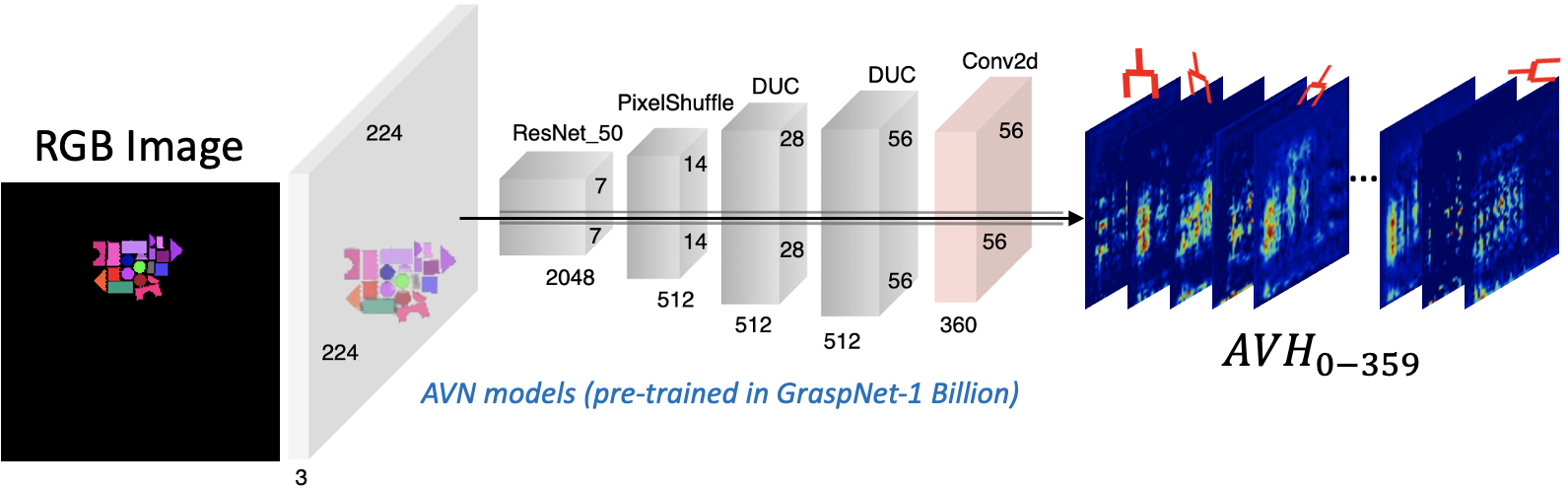}
      \vspace{-4mm}
      \caption{\textbf{Structure of the AVN}: The network takes an RGB image as input, processed by ResNet to extract dense features, then upsampled using pixel shuffle and DUC layers to produce the AVH.}
      \label{fig: avh}
\end{figure}

An example of the AVH is shown in Fig.~\ref{fig: avh}. To map RGB images to the AVH, the AVN uses an encoder-decoder structure. A ResNet50 encoder first transforms the input into high-dimensional features. These are decoded into the AVH through a pixel-shuffle layer and two Dense Upsampling Convolution (DUC) layers, reconstructing spatial orientation information. With view and angle numbers set to 60 and 6, respectively, the final AVH tensor contains 360 feature maps.

\subsubsection{CNN-based RL Model}
\label{subsubsec: cnn}

The AVN maps RGB images to the AVH; however, for our dual-arm push-grasping task, the AVH alone is insufficient due to the complexity of coordinating dual-arm motions. As the AVH is a visual representation, the RL component must be specifically tailored to handle dexterous dual-arm manipulation. Unlike previous methods that rely on DQN and require multiple rotations of RGB-D images to determine action orientation, which leads to complex, heavy models, our approach uses a streamlined CNN-based RL model to process AVN outputs directly. As illustrated in Fig.~\ref{fig: system_overview}, after feature extraction by the AVN, these features are input into our efficient CNN-based RL model.

The policy model is structured with a sequence of $3$ main layers, beginning with a convolutional layer that reduces the input size from $360 \times 56 \times 56$ to $180$ channels, preserving spatial information while reducing computational demands. This is followed by depthwise and pointwise convolution layers, allowing independent channel processing to enhance feature extraction with fewer parameters. The final convolutional layer adjusts the output back to $360$ channels, preparing it for action prediction. A series of $3$ fully connected layers then maps these features into an action vector. In the value model, this convolutional sequence is similarly applied for consistent feature extraction. However, its output is processed by fully connected layers that condense features into a single scalar, representing the expected reward of the state-action pair. The value model provides essential feedback to optimize the policy network. The models guide decision-making, improving performance in complex manipulation tasks.

\subsection{Learning Strategy}
\label{subsec: learning}

\subsubsection{State}
\label{subsubsec: state}

The state of the learning agent is defined by the visual information of the cluttered environment, represented as a 3-channel RGB image. In our framework, this image serves as the observation for the MDP and is sized at $224 \times 224$ pixels. Thus, the state $\bm{s}_t$ at time $t$ is expressed as:

\begin{equation}
    \bm{s}_t = \text{RGB} \in \mathbb{R}^{3 \times 224 \times 224}
    \label{equ: 3}
\end{equation}

\subsubsection{Action}
\label{subsubsec: action}

The action vector of the model is described as a combination of a feature map and the sorted AVH indices, ranked according to predictive scores. The representation of the action output $\bm{a}_t$ can be defined as:

\begin{equation}
    \bm{a}_t = \left[\bm{f}_{\text{map}}, \bm{v}_{\text{AVH}}\right] \in \mathbb{R}^{56 \times 56 + 360}
    \label{equ: 4}
\end{equation}
where, $\bm{f}_{\text{map}}$ represents the feature map of size $56 \times 56 $ derived from the network processing of the RGB input, while $\bm{v}_{\text{AVH}}$ represents the sequence of $360$ sorted AVH indices based on predictive scores. However, the action is designed solely to learn a better feature map representation and cannot be used directly to generate motion. Therefore, we implement two decoders to translate the output into primitive motions.

\begin{figure}[!htbp]
      \centering
      \includegraphics[width=1.0\linewidth]{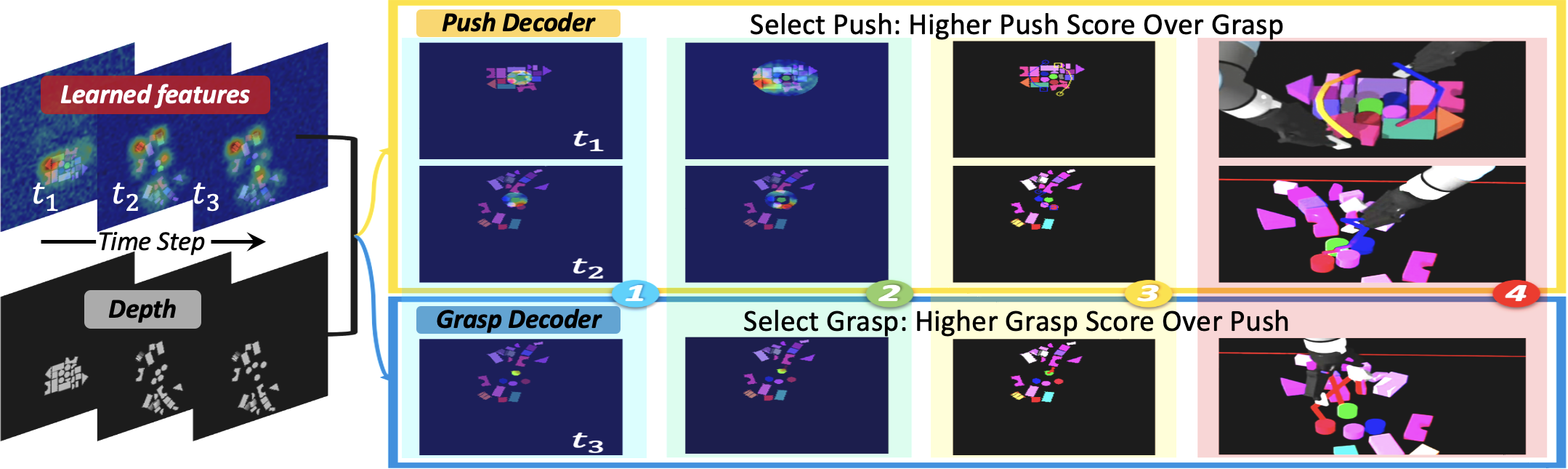}
      \vspace{-4mm}
      \caption{\textbf{Decoders for Dual-Arm Execution:} The decoders translate the feature map into actions \textit{(Sequence of timesteps from a trial)}. A target mask creates the grasp prediction, while an expanded target mask (1.5 times) generates the push score \ccyan{\textit{\textbf{\ding{172}}}}. The action with the higher score, either grasp or push, is selected. For a grasp, the target masked feature map is used; for a push, the mask is adjusted based on the largest contour radius \cgreen{\textit{\textbf{\ding{173}}}}. Grasping extracts translation and orientation; pushing connects key points into paths \corange{\textit{\textbf{\ding{174}}}}. The robot executes the action after motion planning and inverse kinematics \cred{\textit{\textbf{\ding{175}}}}.}
      \label{fig: decoders}
\end{figure}

As shown in Fig.~\ref{fig: decoders}, the decoders convert the learned feature map into executable motions. The target object mask generates a grasp prediction map, while an expanded target mask, scaled $1.5$ times to cover nearby areas, filters the feature map to produce the push prediction score. The action with the highest score, either grasp or push, is then selected. The grasp decoder computes precise grasp configurations, including translation and orientation. Translation is derived from the masked feature map and depth image, while orientation is based on the first angle-view value, as shown in Fig.~\ref{fig: va}. The resulting view and angle values are then used to calculate the final Euler orientation. The push decoder generates paths for rearranging objects in dense clutter. When a push is selected, it calculates the maximum contour radius to create a push mask, connecting key points within this mask into one or two paths. Paths undergo iterative refinement with smoothing and depth validation, ensuring spatial feasibility, followed by 3D smoothing with a Savitzky-Golay filter. To enhance reliability and precision, orientation and safety checks account for workspace constraints, and path planning ensures the starting point is clear of object surfaces for safe pushing. This adaptive approach optimizes push trajectories, providing a robust framework for dual-arm manipulation in complex environments. After final motion planning, safety checks, and inverse kinematics, the robot executes the action.

\subsubsection{Reward}
\label{subsubsec: reward}

We use a fuzzy mechanism-based reward function to guide policy learning for dual-arm pushing and grasping actions. Since our reinforcement learning model focuses on refining feature representations, a consistent evaluation scale is essential for fair and effective learning. To achieve this, the fuzzy reward function provides adaptive feedback at each timestep, enabling nuanced assessments of actions. This approach facilitates learning complex interactions between pushing and grasping motions within a unified framework, enhancing model robustness and accelerating convergence toward optimal manipulation strategies.

\begin{figure}[!t]
      \centering
      \includegraphics[width=1.0\linewidth]{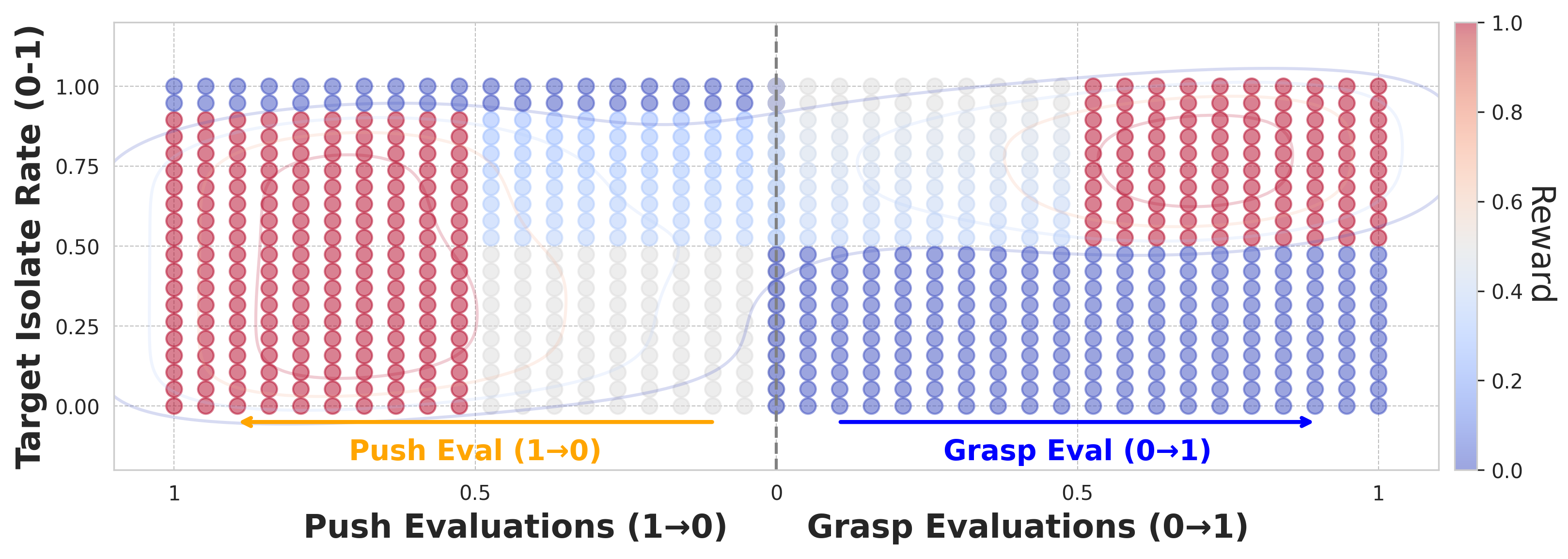}
      \vspace{-4mm}
      \caption{\textbf{Fuzzy Reward}: Optimal actions, like successful grasps at high rates and pushes at low rates, receive a reward of $1.0$ (red). Low-reward actions approach $0.0$ (blue), showing alignment with the target isolate rate.}
      \label{fig: fuzzy_reward}
\end{figure}

The reward function adaptively evaluates dual-arm push-grasp actions, assigning rewards within a range of $0$ to $1$ based on the correctness and success of decisions with target isolate rate levels: very high, medium, or very low. For optimal actions, such as a successful grasp at a high isolate rate or a successful push at a low isolate rate, a very good reward of $1.0$ is assigned. Correct but unsuccessful actions receive a good reward of $0.5$, encouraging progress even without complete success. In medium isolate rate conditions, the reward function applies a nuanced evaluation. Successful actions still receive the maximum reward of $1.0$, while unsuccessful actions are given a neutral reward of $0.25$, adjusted by the target isolate rate. For example, unsuccessful pushes are penalized more heavily as the isolate rate increases, with rewards approaching $0.25$. Conversely, unsuccessful grasps are rewarded more as the isolate rate rises, reflecting their correctness for the target, though still capped below successful actions. Misguided actions, such as a push at a high isolate rate or a grasp at a low isolate rate, are assigned a very bad reward of $0.0$ to discourage decisions that do not align with the intended goal. This adaptive design ensures continuous evaluation, emphasizing correctness and success. By maintaining outputs within $0$ to $1$, it provides strategic feedback to guide decision-making and improve performance, as shown in Fig.~\ref{fig: fuzzy_reward}.


\subsubsection{Policy Optimization}
\label{subsubsec: policy}

We use a single PPO policy to efficiently learn dual-arm pushing and grasping actions, leveraging a fuzzy reward function (Sec.~\ref{subsubsec: reward}) and a CNN-based Actor-Critic architecture (Fig.~\ref{fig: system_overview}). Unlike existing methods with separate reward functions, our approach treats pushing and grasping as interrelated tasks, as both require stable contact with the object and share structural components. This unified reward framework allows the RL model to focus on optimizing feature map outputs for decoding actions, rather than treating each action independently. By viewing pushing and grasping as actions that influence the workspace collectively, the reward function evaluates overall environmental benefit, assigning higher rewards for beneficial actions. This holistic approach enables the agent to learn both actions as a combined process, facilitating efficient learning through a single PPO model and enhancing the synergy between these actions in complex manipulation tasks.

\section{Experiments}
\label{sec: experiments}

In this section, we describe the experimental setup, evaluation metrics, and results of the manipulation tasks conducted in both simulation and real-world environments. Through the experiments, we aim to investigate the following:

\begin{itemize}
    \item Does a single policy network offer superior integration of pushing and grasping actions with greater data efficiency compared to two parallel networks?
    \item Are the primitive dual-arm motions generated by our policy more efficient, enabling tasks to be completed faster and more effectively than with a single-arm robot?
    \item Is the policy robust enough to handle various challenging scenarios, including packed, partial occlusions, complete occlusions, and random densely cluttered environments?
    \item Can the policy transfer to a real robot without fine-tuning and outperform existing push-grasping baselines?
\end{itemize}

\subsection{Experimental Setups}
\label{subsec: experimental}

\begin{figure}[!htbp]
      \centering
      \includegraphics[width=1.0\linewidth]{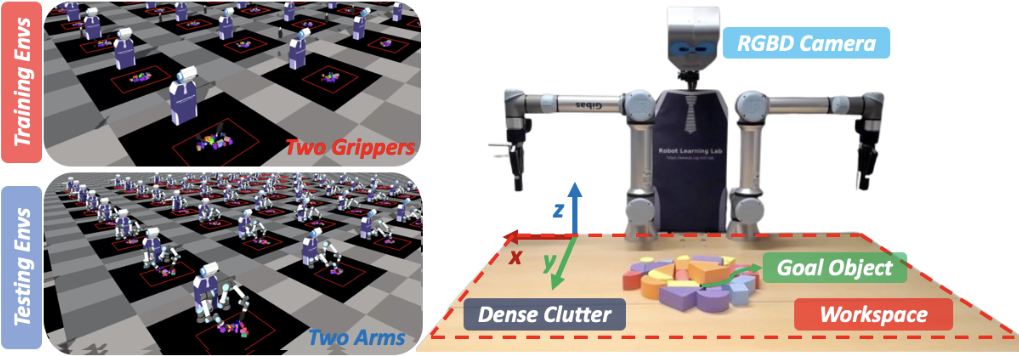}
      \vspace{-4mm}
      \caption{\textbf{Simulation and Real-World Scenarios:} The policy is trained in parallel environments within Isaac Gym, using a simplified setup with two grippers instead of full arms to accelerate training. In real-world testing, the policy is validated on two UR5e robots with Robotiq 2F-140 grippers.}
      \label{fig: exp_envs}
\end{figure}

We trained the proposed system using the Isaac Gym physics simulator and evaluated its performance in both simulated and real-world environments on a dual-arm robotic platform under various cluttered conditions. The simulated and real-world task setups are illustrated in Fig.~\ref{fig: exp_envs}. All experiments were performed on a desktop equipped with two Nvidia RTX 2080Ti GPUs, an Intel i7-9800X CPU, and 12GB of memory allocated per GPU.

\subsection{Evaluation Metrics}
\label{subsec: evaluation}

We evaluate the methods through test cases where the robot rearranges dense clutter to grasp a target object. Each test includes 100 runs in simulation or 10 runs on the real robot, with performance measured by three metrics similar to those in state-of-the-art (SOTA) methods VPG \cite{zeng2018learning}, EPG \cite{xu2021efficient}, and SPG \cite{wang2024self}. All methods are assessed using target-oriented criteria.

\begin{itemize}
    \item \textbf{Completion Rate (C)}: The average percentage of completion across all test runs. A task is considered completed if the policy successfully picks up the target object without experiencing $5$ consecutive failed grasp attempts.
    \item \textbf{Grasp Success Rate (GS)}: The average percentage of successful target object grasps per completed run, reflecting the effectiveness of grasping actions.
    \item \textbf{Motion Number (MN)}: The average number of motions per completion, which indicates action efficiency, particularly in terms of pushing effectiveness. 
\end{itemize}

\subsection{Simulation Experiments}
\label{subsec: simulation}

The simulated environment in Isaac Gym (Fig.~\ref{fig: exp_envs}) captures RGB-D images via a default camera setup. Robots operate in position-control mode, with end-effector positions converted to joint space commands using inverse kinematics. The $1.0 \times 1.0 \, m$ workspace is divided into a $224 \times 224$ grid, each cell corresponding to a pixel in the camera's orthographic image.

\subsubsection{Training Stage}
\label{subsubsec: training}

\begin{figure}[!htbp]
      \centering
      \includegraphics[width=1.0\linewidth]{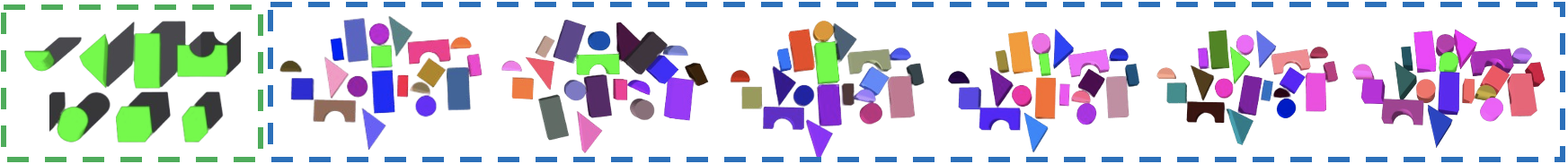}
      \vspace{-4mm}
      \caption{\textbf{Training Scenes}: The scenes are divided into two categories—left (green) for grasp training and right (blue) for push-grasp training.}
      \label{fig: training_stage}
\end{figure}

The training objects are shown in Fig.~\ref{fig: training_stage}. For push-grasping tasks, existing methods typically separate training stages, starting with grasping until a success threshold is reached, then progressing to push-grasping. Models usually use distinct push and grasp modules, making direct push-grasping training rare. Here, we explore whether our model can directly train for push-grasping and if phased training improves performance. In the grasp training phase, no target mask is needed; objects are randomly placed in the workspace. For push-grasping, we use simple cluttered scenes to ensure generalization to more complex settings. We compare our approach to ablation variants to investigate: 1) Does fine-tuning the pre-trained backbone improve performance? 2) Is phased training for pushing and grasping beneficial?

\paragraph{Ablation Experiments}

\begin{figure}[!t]
      \centering
      \includegraphics[width=0.48\linewidth]{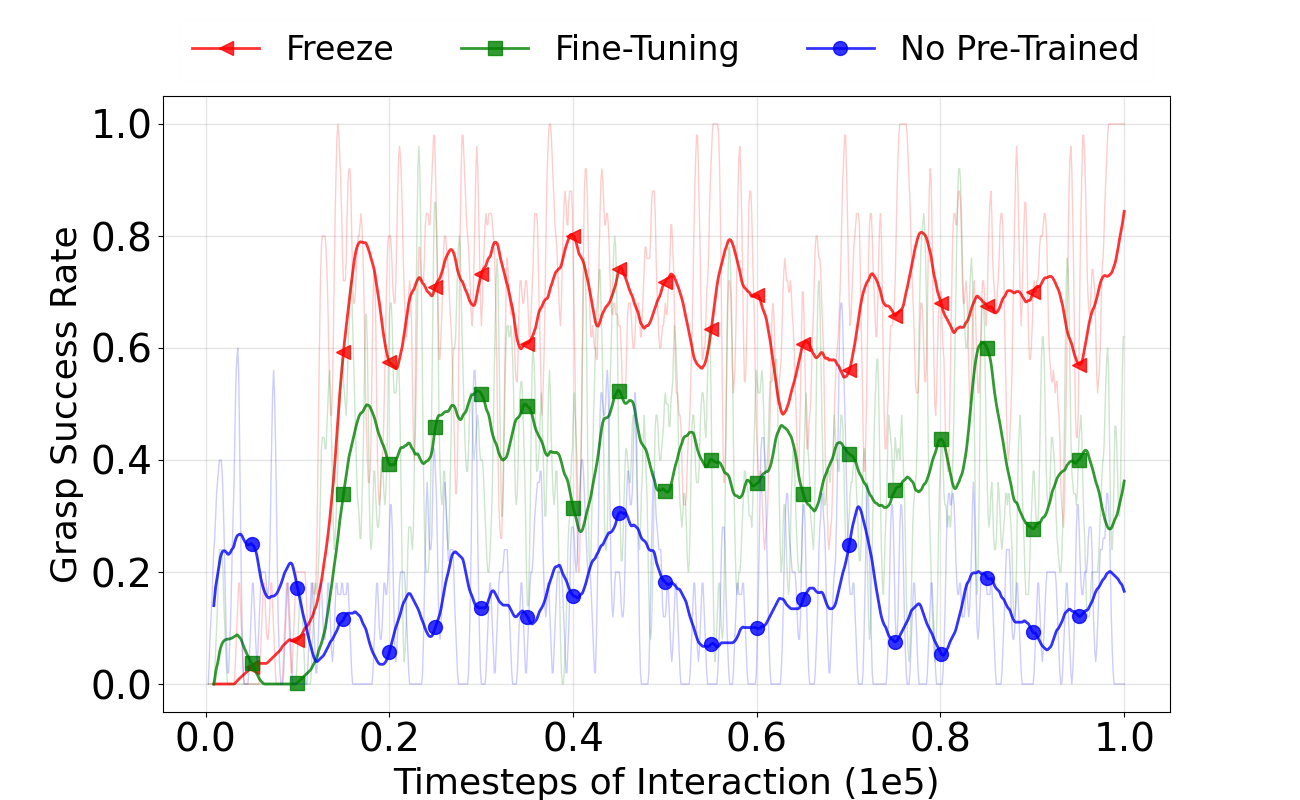}
      \includegraphics[width=0.48\linewidth]{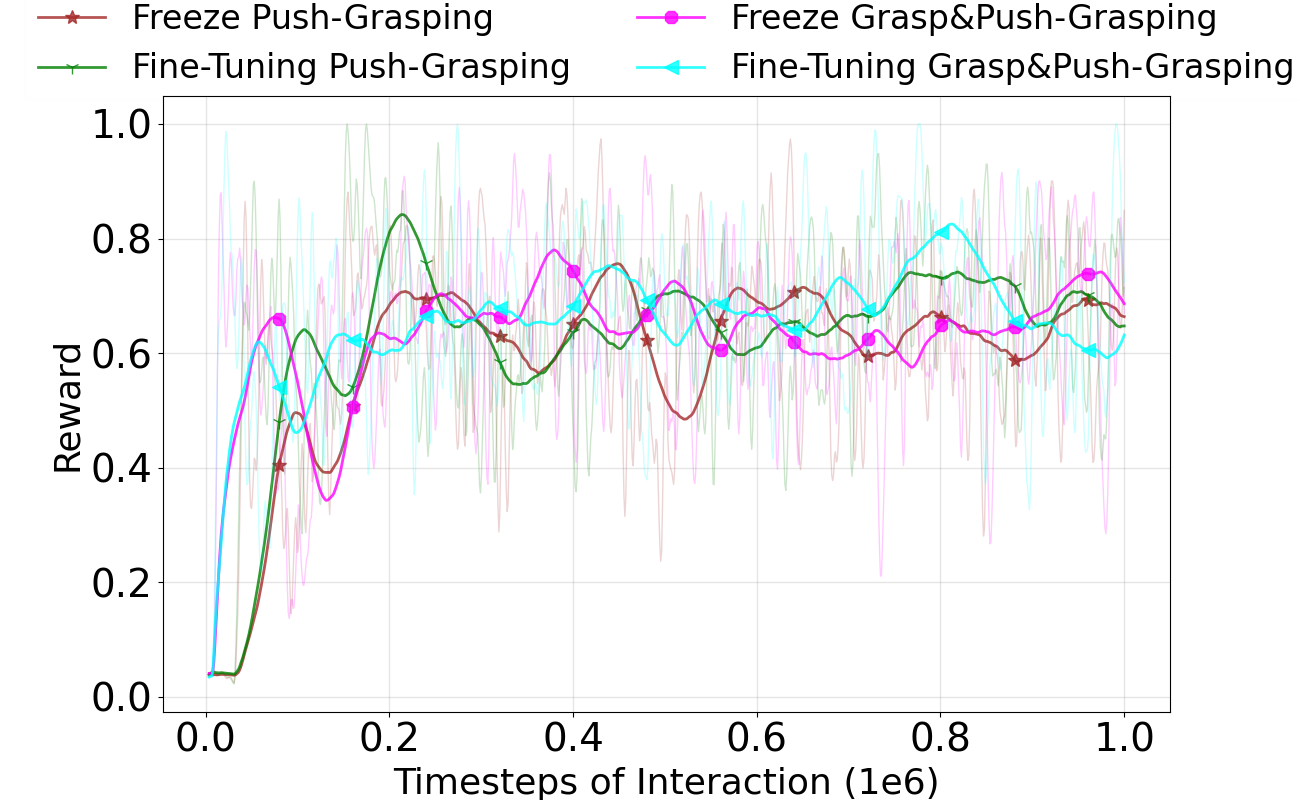}
      \caption{Ablation study to evaluate the training performance of our method, assessing the impact of different training configurations on learning.}
      \vspace{-4mm}
      \label{fig: ablation}

\end{figure}

Our approach employs a large-scale pre-trained model as the backbone for image feature extraction, providing a strong foundation for RL models. A primary question is whether to freeze or fine-tune these pre-trained weights for our task. Additionally, our hierarchical framework's training strategy, especially multi-stage training, requires careful consideration. Results in Fig.~\ref{fig: ablation} indicate that freezing backbone weights outperform updating during training. Furthermore, a two-stage approach (grasping followed by push-grasping) yields slightly better rewards than single-stage push-grasping, suggesting limited benefits of multi-stage training unless objects vary significantly, in which case a separate grasp training phase is advantageous.

\subsubsection{Testing Stage}
\label{subsubsec: testing}

After training, we deployed the model to grasp target objects in clutter, including challenging cases with partial or full occlusions. We created diverse test scenarios with block structures and household objects to evaluate performance on familiar and novel objects.

\paragraph{Comparison Experiments}

\begin{figure*}[!htbp]
      \centering
      \includegraphics[width=1.0\linewidth]{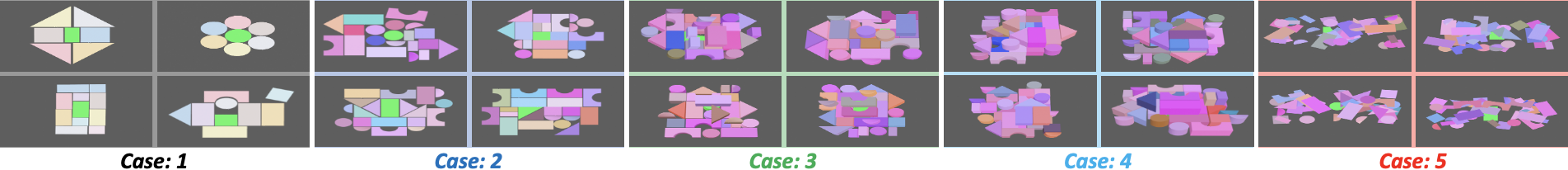}
      \vspace{-4mm}
      \caption{\textbf{Scenes for Comparative Experiments in Simulation}: Challenging simulation scenarios with 5 cases, each consisting of 4 scenes and a unique target object shape. Case 1: packed clutter (up to 10 objects); Case 2: dense clutter (20 objects); Case 3: target partially occluded in dense clutter (25 objects); Case 4: target fully occluded in dense clutter (30 objects); Case 5: target in random dense clutter (35 objects). The target object is marked in green.}
      \label{fig: test_objs_sim}
\end{figure*}

We compared our system with existing baselines. To the best of our knowledge, our method is the first to introduce a dual-arm push-grasping strategy; therefore, we benchmarked it against state-of-the-art single-arm methods. We ran 5 rounds of experiments, each with 4 distinct scenes. All table data represents averages across scenes; detailed metrics are in the supplementary materials. While our method (single-arm) performs comparably in some clutter scenes, it falls short in most dense clutter scenes. The choice between single- and dual-arm setups affects scene evolution, influencing task success.

\textbf{Clutter:} In the first round, $4$ differently shaped objects are used in a packed clutter scenario with increasing complexity, as shown in Fig.~\ref{fig: test_objs_sim}, Case 1. As indicated in Table~\ref{table: sim_results_1} Case 1, our method completes the task with fewer motions and a higher success rate compared to other baselines. 

\textbf{Dense Clutter:} In the second round, the target object is surrounded by 20 objects in a dense cluttered environment, as shown in Fig.~\ref{fig: test_objs_sim}, Case 2. Table~\ref{table: sim_results_1}, Case 2, demonstrates that our method achieves significantly fewer motions and a higher success rate than other baselines. This improvement is not solely due to the dual-arm setup; rather, the model effectively learns key points from images, coupled with a well-designed action strategy, resulting in the highest grasp success rate.

\textbf{Partially Occluded Clutter:} In the third round, the target object is partially occluded among 25 objects in a dense scene, as shown in Fig.~\ref{fig: test_objs_sim}, Case 3. Table~\ref{table: sim_results_2}, Case 3, shows that despite the high object count, our method maintains efficiency.

\textbf{Completely Occluded Clutter:} In the fourth round, the target object is fully occluded among 30 objects, as shown in Fig.~\ref{fig: test_objs_sim}, Case 4. Table~\ref{table: sim_results_2}, Case 4, shows our method achieves a higher grasp success rate with fewer motions. Unlike methods that rely on thresholds for grasp suitability, often defaulting to a push when quality is low or border occupancy is high, our approach learns decisions directly from images. Our grasps are not restricted to top-down poses, enhancing effectiveness in dense clutter, and our push actions are longer and more adaptable, helping isolate the target object.

\textbf{Randomly Arranged Clutter:} In the fifth round, 35 objects are randomly placed in the workspace, as shown in Fig.~\ref{fig: test_objs_sim}, Case 5. Table~\ref{table: sim_results_2}, Case 5, shows our method outperforms others, though with a smaller performance gap than in previous cases due to the randomness of placements. The target may be accessible or fully occluded, but our method remains efficient.

\begin{table}[!t]
    \centering
    \caption{Simulation Results on Case:1-2 Scenes (100 Trials per Scene)}
    \renewcommand{\arraystretch}{1.1} 
    \resizebox{\linewidth}{!}{
    \begin{tabular}{ccccc}
    \hline
    \textbf{Approach} & \textbf{Scene ID} & \textbf{C\%} & \textbf{GS\%} & \textbf{MN} \\
    \hline
         & Case:1 & $91.50 \pm 2.76$ & $47.76 \pm 3.42$ & $7.27 \pm 0.35$ \\
    VPG  & \cblue{Case:2} & $83.00 \pm 3.76$  & $44.91 \pm 3.23$ & $13.85 \pm 1.44$ \\
    \hline
         & Case:1 & $96.00 \pm 1.62$  & $58.55 \pm 3.80$ & $6.48 \pm 0.28$ \\
    EPG  & \cblue{Case:2} & $87.75 \pm 3.14$ & $52.07 \pm 3.57$ & $13.41 \pm 1.15$ \\ 
    \hline
         & Case:1 & $98.75 \pm 0.95$  & $69.90 \pm 3.27$ & $6.07 \pm 0.40$ \\
    SPG  & \cblue{Case:2} & $89.50 \pm 3.02$ & $57.22 \pm 3.74$ & $11.60 \pm 0.79$ \\ 
    \hline
         & Case:1 & $99.25 \pm 0.60$  & $90.55 \pm 2.86$ & $2.93 \pm 0.11$ \\
    Ours (single-arm)  & \cblue{Case:2} & $98.00 \pm 1.35$ & $81.34 \pm 3.62$ & $4.74 \pm 0.25$ \\ 
    \hline
         & Case:1 & $\mathbf{99.75 \pm 0.25}$ & $\mathbf{93.08 \pm 2.46}$ & $\mathbf{2.79 \pm 0.12}$ \\
    Ours & \cblue{Case:2} & $\mathbf{98.51 \pm 1.20}$ & $\mathbf{83.70 \pm 3.49}$ & $\mathbf{4.05 \pm 0.18}$ \\ 
    \hline
    \end{tabular}}
    \label{table: sim_results_1}
\end{table}

\begin{table}[!t]
    \centering
    \caption{Simulation Results on Case:3-5 Scenes (100 Trials per Scene)}
    \renewcommand{\arraystretch}{1.1} 
    \resizebox{\linewidth}{!}{
    \begin{tabular}{ccccc}
    \hline
    \textbf{Approach} & \textbf{Scene ID} & \textbf{C\%} & \textbf{GS\%} & \textbf{MN} \\
    \hline
         & \cgreen{Case:3}  & $84.50 \pm 3.54$   & $50.26 \pm 3.36$ & $16.13 \pm 1.73$ \\
    EPG  & \ccyan{Case:4} & $85.00 \pm 3.35$ & $47.98 \pm 3.34$ & $30.92 \pm 4.60$ \\
         & \cred{Case:5} & $91.25 \pm 2.50$ & $55.58 \pm 3.58$ & $13.03 \pm 0.80$ \\ 
    \hline
         & \cgreen{Case:3}  & $86.00 \pm 3.35$   & $58.86 \pm 3.68$ & $16.83 \pm 1.82$ \\
    SPG  & \ccyan{Case:4} & $91.00 \pm 2.77$ & $52.98 \pm 3.61$ & $29.96 \pm 1.62$ \\
         & \cred{Case:5} & $97.25 \pm 1.60$ & $67.17 \pm 3.88$ & $14.40 \pm 0.64$ \\
    \hline
         & \cgreen{Case:3}  & $98.25 \pm 0.88$   & $83.77 \pm 3.77$ & $\mathbf{5.64 \pm 0.18}$ \\
    Ours (single-arm)  & \ccyan{Case:4} & $96.50 \pm 1.82$ & $79.42 \pm 3.78$ & $6.48 \pm 0.34$ \\
         & \cred{Case:5} & $\mathbf{98.80 \pm 0.92}$ & $73.07 \pm 3.75$ & $\mathbf{5.05 \pm 0.22}$ \\
    \hline
         & \cgreen{Case:3} & $\mathbf{98.50 \pm 1.03}$ & $\mathbf{84.64 \pm 3.42}$ & $5.71 \pm 0.31$ \\
    Ours & \ccyan{Case:4} & $\mathbf{97.00 \pm 1.70}$ & $\mathbf{81.25 \pm 3.70}$ & $\mathbf{6.09 \pm 0.32}$ \\ 
         & \cred{Case:5} & $98.75 \pm 0.93$ & $\mathbf{78.49 \pm 3.69}$ & $6.05 \pm 0.30$ \\
    \hline
    \end{tabular}}
    \label{table: sim_results_2}
\end{table}

\subsection{Real-World Experiments}
\label{subsec: real}

\begin{figure*}[!htbp]
      \centering
      \includegraphics[width=1.0\linewidth]{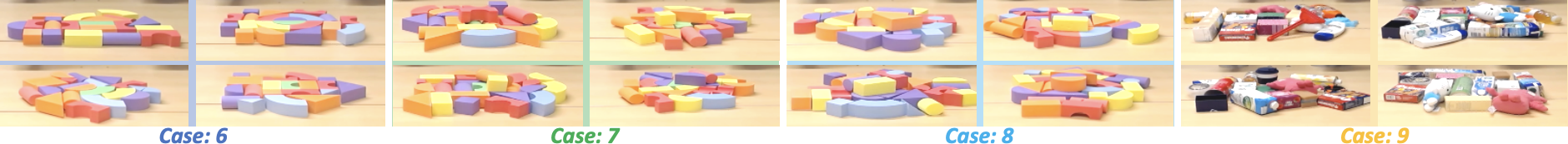}
      \vspace{-4mm}
      \caption{\textbf{Real-World Scenes for Comparative Experiments:} Challenging real-world scenarios with 4 cases, each featuring 4 scenes and a distinct target object shape. Case 6: dense clutter (20 objects); Case 7: target partially occluded within dense clutter (25 objects); Case 8: target fully occluded within dense clutter (30 objects); Case 9: target surrounded by various household items in dense clutter (15 objects). The target object is highlighted in green.}
      \label{fig: test_objs_real}
\end{figure*}

We evaluated our system in real-world experiments. The setup is shown in Fig.~\ref{fig: exp_envs}, using RGB-D images captured by an Asus Xtion camera. As illustrated in Fig.~\ref{fig: test_objs_real}, our test cases include 12 scenes with building blocks and 4 with household objects. For building blocks, tests include 4 dense clutter scenarios, 4 with partial occlusion, and 4 with full occlusion of the target. For household objects, all 4 scenarios involve dense clutter. Notably, our models were transferred directly from simulation to the real world without retraining. Results in Table~\ref{table: real_robot_results} show that our policy generalizes effectively to real-world settings and adapts to some unseen objects. 


\begin{table}[!t]
    \centering
    \caption{Real Robot Results on All Scenes (10 Trials per Scene)}
    \renewcommand{\arraystretch}{1.1} 
    \resizebox{\linewidth}{!}{
    \begin{tabular}{ccccc}
    \hline
    \textbf{Approach} & \textbf{Scene ID} & \textbf{C\%} & \textbf{GS\%} & \textbf{MN} \\
    \hline
         & \cblue{Case:6}  & $90.00 \pm 8.33$   & $54.58 \pm 11.82$ & $14.23 \pm 1.71$ \\
    SPG  & \cgreen{Case:7} & $85.00 \pm 11.67$  & $48.48 \pm 10.82$ & $20.65 \pm 3.33$ \\ 
         & \ccyan{Case:8}  & $85.00 \pm 11.64$  & $48.83 \pm 10.92$ & $30.13 \pm 3.33$ \\ 
         & \corange{Case:9} & $95.00 \pm 5.00$   & $64.68 \pm 12.33$ & $10.08 \pm 1.28$ \\ 
    \hline
         & \cblue{Case:6}  & $\mathbf{97.50 \pm 4.77}$ & $\mathbf{81.91 \pm 9.17}$ & $\mathbf{5.43 \pm 0.60}$ \\
    Ours & \cgreen{Case:7} & $\mathbf{95.00 \pm 5.00}$  & $\mathbf{74.25 \pm 11.79}$ & $\mathbf{6.75 \pm 1.61}$ \\ 
         & \ccyan{Case:8}  & $\mathbf{92.50 \pm 7.50}$  & $\mathbf{68.87 \pm 12.48}$ & $\mathbf{7.53 \pm 1.28}$ \\ 
         & \corange{Case:9} & $\mathbf{97.50 \pm 2.50}$  & $\mathbf{78.42 \pm 11.33}$ & $\mathbf{4.78 \pm 0.63}$ \\ 
    \hline
    \end{tabular}}
    \label{table: real_robot_results}
\end{table}

\subsection{Failure Cases}


\textbf{Out-of-Bounds Object Displacement:} An experiment fails when an object leaves the workspace.

\textbf{Unreachable Push Path Points:} Despite safety checks, push paths sometimes include unreachable points.

\textbf{Imprecise Grasp:} Failures from grasp pose errors led to missed or unintended multi-object grasps.

\section{Conclusion}
\label{sec: conclusion}

In this paper, we present a visual-based dual-arm push-and-grasp approach for densely cluttered environments, where grasping a target object requires first pushing obstructions aside. To tackle this task efficiently, we use a large-scale grasp model as the backbone for feature extraction from images and develop a CNN-based PPO model to learn an optimal dual-arm push-grasp strategy. We rigorously evaluate our method in diverse, challenging scenarios in both simulation and real-world settings. Simulation results demonstrate that our approach effectively trains a policy adaptable to complex scenarios, achieving high performance in random and dense clutter cases. The trained policy transfers seamlessly to real-world applications without fine-tuning, underscoring its robustness and generalization capabilities. Our model enables dual-arm robots to perform coordinated push-grasp motions, efficiently handling dense clutter with both arms. For future work, we plan to expand our framework for complex scenarios and integrate open-world grasping with large vision-language models \cite{tziafas2024OWG} to advance real-world autonomous manipulation.

\bibliographystyle{IEEEtran}
\bibliography{reference}

\end{document}